\def\year{2019}\relax
\documentclass[letterpaper]{article} 
\usepackage{aaai19}  
\usepackage{graphicx}
\usepackage{amsmath}
\usepackage{times}  
\usepackage{helvet}  
\usepackage{courier}  
\usepackage{url}  
\usepackage{graphicx}  
 \usepackage{mathrsfs}
 \usepackage{multirow}
\newenvironment{myitemize2}[1][]{
\begin{list}{$\bullet$}
    {
     \setlength{\leftmargin}{5mm}     
     \setlength{\parsep}{0.5mm}         
     \setlength{\topsep}{0mm}         
     \setlength{\itemsep}{0mm}        
     \setlength{\labelsep}{0.5em}     
     \setlength{\itemindent}{0mm}    
     \setlength{\listparindent}{6mm} 
    }}
{\end{list}}

\begin{document}
%
\title{A Deep Cascade Model for Multi-Document Reading Comprehension}

\author{Ming Yan, Jiangnan Xia, Chen Wu, Bin Bi\\  \bf \Large  Zhongzhou Zhao, Ji Zhang, Luo Si, Rui Wang, Wei Wang, Haiqing Chen 
\\ Alibaba Group
\\ \{ym119608, jiangnan.xjn, wuchen.wc, b,bi\}@alibaba-inc.com
 \\ \{zhongzhou.zhaozz, zj122146, luo.si, masi.wr, hebian.ww, haiqing.chenhq\}@alibaba-inc.com} 

\maketitle
\begin{abstract}
A fundamental trade-off between effectiveness and efficiency needs to be balanced when designing an online question answering system. Effectiveness comes from sophisticated functions such as extractive machine reading comprehension (MRC), while efficiency is obtained from improvements in preliminary retrieval components such as candidate document selection and paragraph ranking. Given the complexity of the real-world multi-document MRC scenario, it is difficult to jointly optimize both in an end-to-end system. To address this problem, we develop a novel deep cascade learning model, which progressively evolves from the document-level and paragraph-level ranking of candidate texts to more precise answer extraction with machine reading comprehension. Specifically, irrelevant documents and paragraphs are first filtered out with simple functions for efficiency consideration. Then we jointly train three modules on the remaining texts for better tracking the answer: the document extraction,  the paragraph extraction and the answer extraction. Experiment results show that the proposed method outperforms the previous state-of-the-art methods on two large-scale multi-document benchmark datasets, i.e., TriviaQA and DuReader. In addition, our online system can stably serve typical scenarios with millions of daily requests in less than 50ms.
\end{abstract}

\section{Introduction}

Machine reading comprehension (MRC), which empowers computers with the ability to read and comprehend knowledge and then answer questions from textual data, has made rapid progress in recent years. From the early cloze-style test~\cite{hermann2015teaching,hill2015goldilocks} to answer extraction from a single paragraph~\cite{rajpurkar2016squad}, and to the more complex open-domain question answering from web data~\cite{joshi2017triviaqa,nguyen2016ms}, great efforts have been made to push the MRC technique to more practical applications.

The rapid progress of MRC in recent years mostly owes to the release of the single-paragraph benchmark dataset SQuAD~\cite{rajpurkar2016squad}, on which various deep attention-based methods have been proposed to constantly push the state-of-the-art performance~\cite{seo2016bidirectional,wang2017gated,yu2018qanet}. It is a significant milestone that several MRC models have exceeded the performance of human annotators on the SQuAD dataset\footnote{\small{~https://rajpurkar.github.io/SQuAD-explorer/}}. However, the SQuAD dataset makes a strong assumption that the answers are contained in the given paragraphs. Besides, the parapraphs are rather short, approximately 200 words on average, while a real-world scenario usually involves multiple documents of much longer length. Therefore, several latest studies~\cite{joshi2017triviaqa,clark2017simple,tan2017s} begin to re-design the task into more realistic settings: the MRC models are required to read and comprehend multiple documents to reach the final answer.

In multi-document MRC, depending on the way of combining the two components, document selection and extractive reading comprehension, there are two categories of approaches: 1) The pipeline approach treats the document selection and extractive reading comprehension as two separate parts, where a document is firstly selected through document ranking and then passed to the MRC model for extracting the final answer~\cite{joshi2017triviaqa,wang2017r}; 2) Several recent studies~\cite{tan2017s,clark2017simple,wang2018multi} adopt a joint learning method to optimize both sub-tasks in a unified framework simultaneously.

The pipeline method relies heavily on the quality of the document ranking module. When it fails to give the relevant documents higher ranks or filters out the ones that contain the correct answers, the downstream MRC module has no way to recover and extract the answers of interest. For the joint learning method, it is computationally expensive to jointly optimize both tasks with all the documents. This computation cost limits its application to the operational online environment, such as Amazon\footnote{\small{https://www.amazon.com/}} and Taobao\footnote{\small{https://www.taobao.com/}}, where efficiency is a critical factor to be considered. 

To address the above problems, we propose a deep cascade model which combines the advantages of both methods in a coarse-to-fine manner. The deep cascade model is designed to properly keep the balance between the effectiveness and efficiency. At early stages of the model, simple features and ranking functions are used to select a candidate set of most relevant contents, filtering out the irrelevant documents and paragraphs as much as possible. Then the selected paragraphs are passed to the attention-based deep MRC model for extracting the actual answer span at word level. To better support the answer extraction, we also introduce the document extraction and paragraph extraction as two auxiliary tasks, which helps to quickly narrow down the entire search space. We jointly optimize all the three tasks in a unified deep MRC model, which shares some common bottom layers. This cascaded structure enables the models to perform a coarse-to-fine pruning at different stages, better models can be learnt effectively and efficiently.



\begin{figure}
\centering
\includegraphics[width=0.465\textwidth]{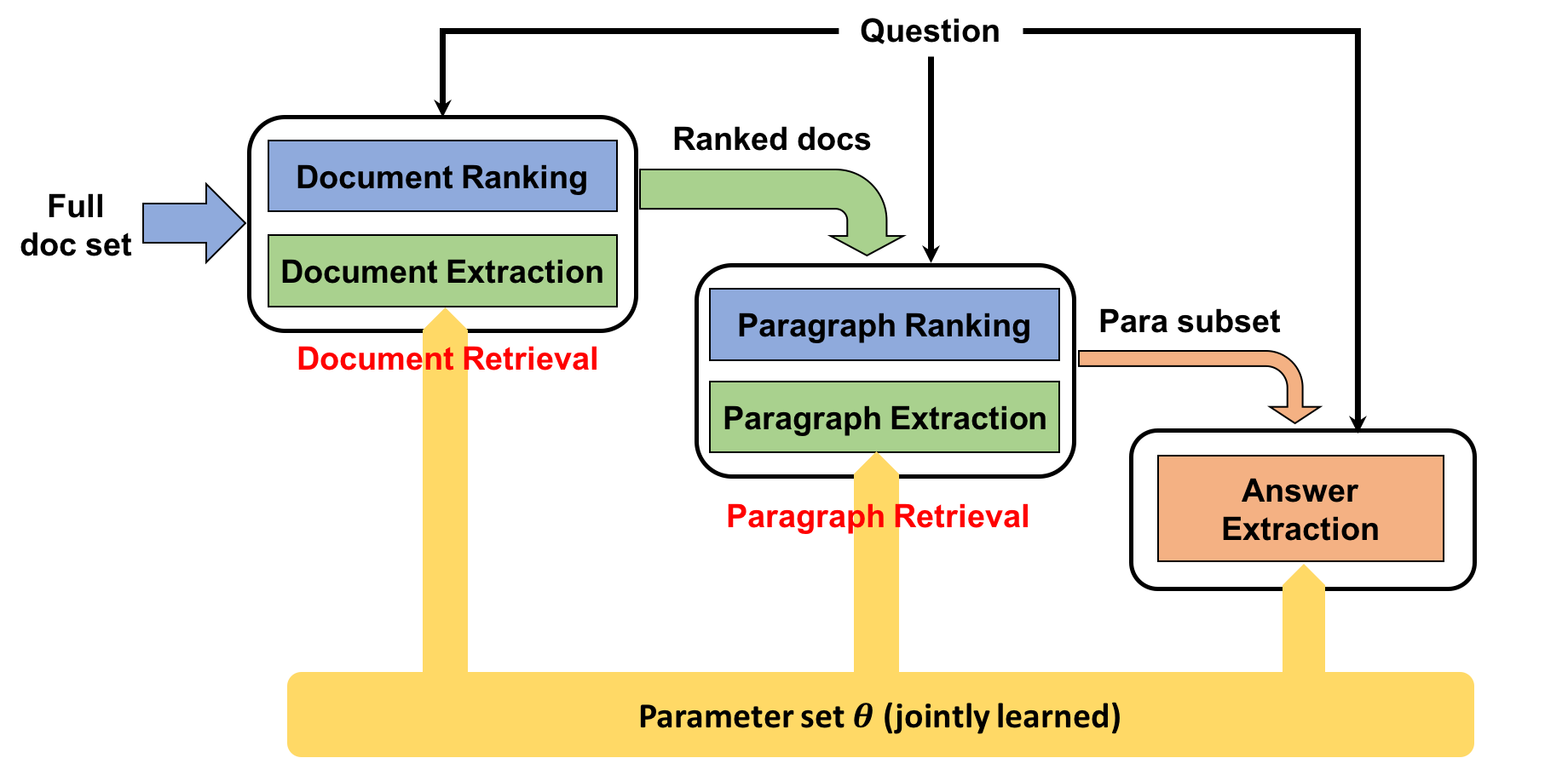}\vspace{-2mm}
\caption{The overall framework of our deep cascade model, which consists of the document retrieval, paragraph retrieval and answer extraction modules.} \vspace{-3mm}
\label{fig:arch}
\end{figure}

The overall framework of our model is demonstrated in Figure~\ref{fig:arch}, which consists of three modules: document retrieval, paragraph retrieval and answer extraction. The first  module takes the question and a collection of raw documents as input. The module at each subsequent stage consumes the output from the previous stage, and further prunes the documents, paragraphs and answer spans given the question. For each of the first two modules, we define a ranking function and an extraction function. The ranking function is first used as a preliminary filter to discard  most of the irrelevant documents or paragraphs, so as to keep our framework efficient. The extraction function is then designed to deal with the auxiliary document and paragraph extraction tasks, which is jointly optimized with the final answer extraction module for better extraction performance.  The local ranking functions in different modules gradually increase in cost and complexity, to properly keep the balance between the effectiveness and efficiency.


The main contributions can be summarized as follow:
\begin{myitemize2}
	\item We propose a deep cascade learning framework to address the practical multi-document machine reading comprehension task,  which considers both the effectiveness and efficiency in a coarse-to-fine manner.
	\item We incorporate the auxiliary document extraction and paragraph extraction tasks to the pure answer span prediction, which helps to narrow down the search space and improves the final extraction result in multi-document MRC scenario.
	\item We conduct extensive experiments on two large-scale multi-document MRC benchmark datasets: TriviaQA~\cite{joshi2017triviaqa} and DuReader~\cite{he2017dureader}. The results show that our deep cascade model can outperform the previous state-of-the-art performance on both datasets. Besides,  the proposed model has also been successfully applied in our online system and stably serve various scenarios in a quick response time of less than 50ms.
\end{myitemize2}


\section{Related Work}

\subsection{Machine Reading Comprehension}

Recently, we can see emerging interests in multi-document MRC research~\cite{nguyen2016ms,clark2017simple,wang2017evidence,he2017dureader,wang2018multi}, where multiple documents are given as input. There are two categories of approaches: the pipeline-based approaches and the joint learning models. The pipeline approach firstly selects a single document via ranking and then pass it to the MRC model to extract the precise answer~\cite{joshi2017triviaqa,wang2017r}.
This approach gives huge burden to the document ranking model, in which the downstream MRC model has no way to extract the right answer if the relevant documents are missed. The joint learning approaches take all the documents into consideration and extract the answer by comparing it against other documents~\cite{clark2017simple,tan2018s,wang2018multi}.
\cite{clark2017simple} proposes a confidence-based method with a shared normalization training objective, which enables the model to produce globally correct output. \cite{tan2018s} proposes an extraction-then-synthesis framework, by also incorporating passage ranking to answer span prediction. \cite{wang2018multi} further proposes a verification method to make use of the extracted answers in different documents to verify each other for more accurate prediction. However, taking all the documents into consideration will inevitably bring more computation cost, which can be unbearable in the operational online environment. Our deep cascade model can serve as a proper tradeoff between the pipeline method and joint learning method. It has a coarse-to-fine structure which can eliminate irrelevant documents and paragraphs in the early stages with simple features and models, and better identify more relevant answers in a well-designed multi-task deep MRC model on the remaining content.

\subsection{Cascade Learning}
In designing online systems, trade-off between effectiveness and efficiency remains a long-standing problem. Cascade learning is an alternative strategy that can better balance these two, which utilizes a sequence of functions in different stages and allows using different sets of features for different instances. It is firstly introduced in the traditional classification and detection problems such as fast visual object detection~\cite{schneiderman2004feature,bourdev2005robust}, and then widely applied in ranking applications for achieving high top-k rank effectiveness in an efficient manner~\cite{lefakis2010joint,wang2011cascade,liu2017cascade}. \cite{wang2011cascade} uses an Adaboost style framework with two independent ranking functions in each stage, one for pruning the input ranked documents and the other for refining the rank order.

We apply the idea of cascade learning to machine reading comprehension, from a preliminary document-level and paragraph-level ranking of the candidate texts, to a more precise answer span extraction. The extracted answer spans are progressively narrowed down across different levels, and the ranking and extraction functions also progressively increase in complexity for more precise answer prediction. 

\section{The Deep Cascade Model}

\begin{figure}
\centering
\includegraphics[width=0.48\textwidth]{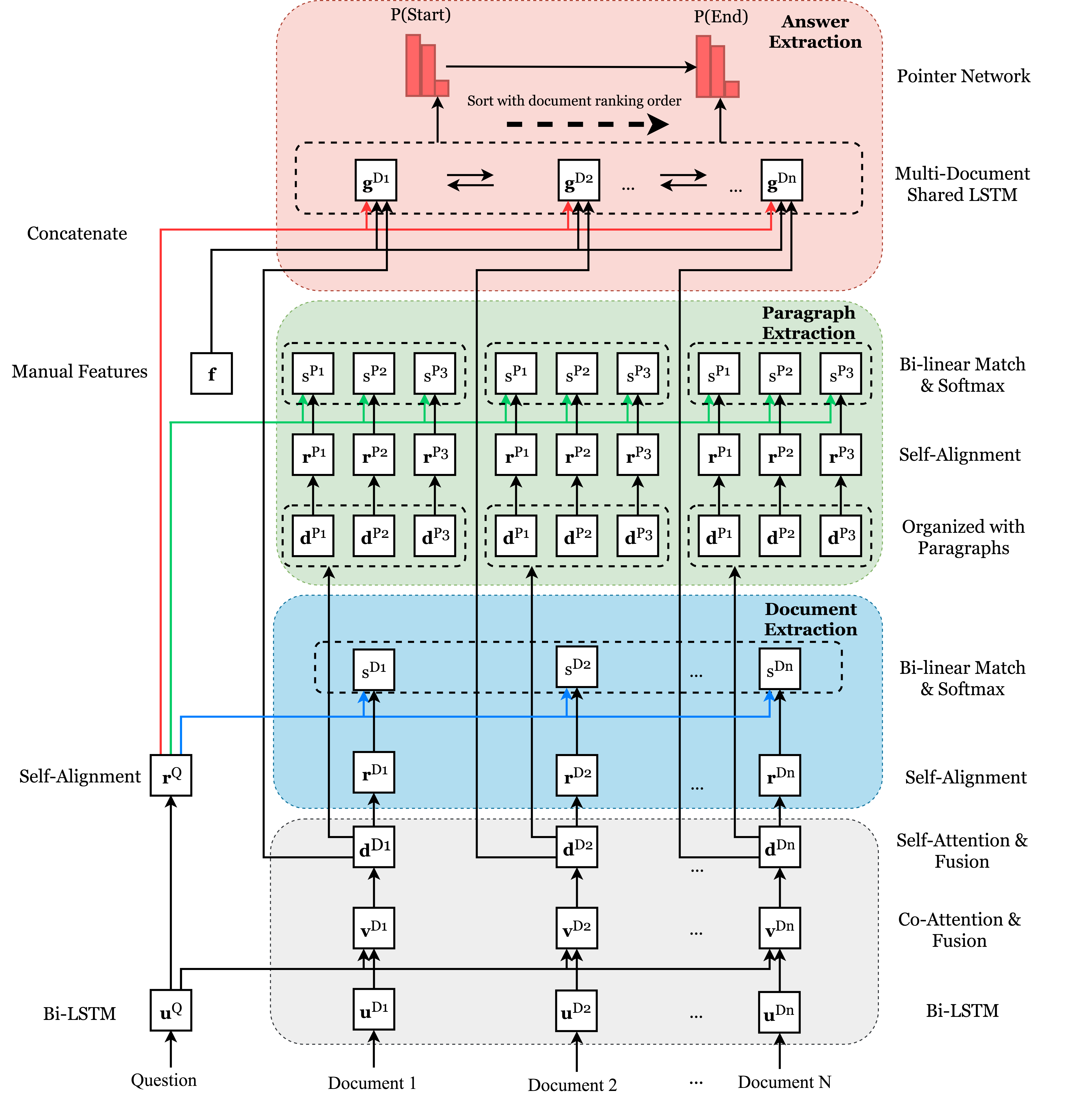}\vspace{-2mm}
\caption{The deep attention-based multi-task MRC model.}\vspace{-3mm}
\label{fig:cascade}
\end{figure}

Following the overview in Figure~\ref{fig:arch}, our approach consists of three cascade modules: document retrieval, paragraph retrieval and answer extraction. The cascade ranking functions in the first two modules aim to fast filter out the irrelevant document content based on the basic statistical and structural features, and obtain a coarse ranking for the candidate documents. For the remaining document content, we design three extraction tasks at different granularities, with the goal to simultaneously extract the right document, paragraph and also the answer span. A deep attention-based MRC model is designed to jointly optimize all the three extraction tasks, by sharing the common bottom layers, as is shown in Figure~\ref{fig:cascade}. The final answer is thus determined by not only the answer span prediction score, but also the corresponding document and paragraph prediction score. 

\subsection{Cascade Ranking Functions}
Given a question $\textbf{Q}$ and a set of candidate documents \{$\textbf{D}_i$\}, we first introduce the cascade ranking functions of the first two modules for pruning the documents, which gradually increases in complexity.  

\subsubsection{Document Ranking}
This part aims at fast filtering out the irrelevant documents and obtaining a coarse ranking for the candidate documents. We utilize the traditional information retrieval methods, such as BM25 and TF-IDF distance, to measure the relevance between the question and document. The matching is conducted between the textual metadata of question and document, including the document title and main content. Besides, the recall ratio of the question words from the document metadata is used as another feature to indicate the relevance of the document. 

To learn the importance of different features, we use a learning-to-rank model to assign a weighted relevance score to each retrieved document. By design, the first stage needs to be quick and simple, so we cast the task as a binary classification problem and adopt the pointwise logistic regression as the ranking function. The documents containing the answer are labeled as positive. After this ranking, we only keep the top-K ranked documents for further processing. 

\subsubsection{Paragraph Ranking}
This part aims at fast discarding the irrelevant content within each document at a paragraph level. Specifically, given an output document $\textbf{D}_i=\{\textbf{P}_{ij}\}$ from the previous stage, we first prune the noisy paragraphs without word or entity match. The simple question and paragraph textual matching features are also extracted as in that of document ranking. Moreover, the document structure can contain some sort of inherent information, for example, the first paragraph within a document may tend to possess more informative content as a document abstract. Therefore, we also add some structural features, such as whether the paragraph is the first or last paragraph of the document, the length of the paragraph, the length of the previous or subsequent paragraphs. To understand the question, we also incorporate the question type information as several binary features if is given, e.g. for DuReader dataset. 

To better combine different kinds of features, we adopt a scalable tree boosting method XGBoost~\cite{chen2016xgboost} for ranking, which is widely used to achieve state-of-the-art results on many large-scale machine learning challenges. Again, we use the binary logistic loss for model training and label the paragraph containing the answer as positive. As a result, we select the top-N paragraphs from each document for the subsequent answer prediction. 

\subsection{Multi-task Deep Attention Model}
Given the selected P paragraphs from the top-ranked K documents, the final task is to extract an answer span to answer the question $\textbf{Q}$. A deep attention-based MRC model is designed to achieve this goal. However, with all these documents and paragraphs, it may be still difficult to directly conduct the pure answer prediction at a precise word level, as in that of SQuAD dataset. The document and paragraph information is also not fully exploited. Therefore, we split the answer prediction task into three joint tasks: document extraction, paragraph extraction and answer span extraction. The three tasks share the same bottom layers, which represents the semantics of the document context with respect to the question words, as is shown in Figure~\ref{fig:cascade}. By introducing the auxiliary document extraction and paragraph extraction tasks, the proposed model can progressively narrow down the search space from coarse to fine, which helps to better locate on the final answer span. The final answer prediction is based on the results of all the three tasks, which is jointly optimized with a joint learning method. 

\subsubsection{Shared Q\&D Modeling}
Given a question $\textbf{Q}$ and a set of selected documents \{$\textbf{D}_i$\}, one of the keys in MRC model lies in how to incorporate the question context into the document, so that important information can be highlighted. We follow the attention \& fusion mechanism used in \cite{wang2018alibaba}, which is a previous state-of-the-art MRC method on SQuAD dataset.  

Specifically, we first map each word into the vector space by concatenating its word embedding and CNN-based character embedding. Then we use bi-directional LSTM (BiLSTM) to encode the question $\textbf{Q}$ and documents \{$\textbf{D}_i$\} as:
\begin{equation}\label{equ:2}
\begin{split}
&\textbf{u}_t^Q =  {\rm BiLSTM}_{Q}(\textbf{u}_{t-1}^Q, [\textbf{e}_t^Q,\textbf{c}_t^Q])  \\
&\textbf{u}_t^D =  {\rm BiLSTM}_{D}(\textbf{u}_{t-1}^D, [\textbf{e}_t^D,\textbf{c}_t^D])
\end{split}
\end{equation}
where $\textbf{e}_t$ and $\textbf{c}_t$ are the word embedding and character embedding of the $t^{th}$ word. $\textbf{u}_t^Q$ and $\textbf{u}_t^D$ are the encoding vectors of the $t^{th}$ word in $\textbf{Q}$ and $\textbf{D}$, respectively.

After the encoding, we use the co-attention method to effectively incorporate the question information into the document context, and obtain the question-aware document representation $\tilde{\textbf{u}}_t^D=\sum_{j}\alpha_{tj}\cdot \textbf{u}_j^D$. We adopt the attention function used in DrQA~\cite{chen2017reading}, which computes the attention score $\alpha_{ij}$ by the dot products between nonlinear mappings of word representations:
\begin{equation}\label{equ:3}
 \alpha_{ij} =\rm{softmax}({\rm ReLU}(W^\top_{l} \textbf{u}_i^Q)^\top \cdot {\rm ReLU}(W^\top_{l} \textbf{u}_j^D)) \\
\end{equation}
where $W_{l}$ is a linear projection matrix, $\rm softmax$ is the normalization function, and $\rm ReLU$ is the nonlinear activation function. 

To combine the original representation $\textbf{u}_t^D$ and the attention vector $\tilde{\textbf{u}}_t^D$, we adopt the fusion kernel used in~\cite{wang2018alibaba} for better semantic understanding: 
\begin{equation} \label{equ:4}
  \textbf{v}_t^D={\rm Fuse}(\textbf{u}_t^D,\tilde{\textbf{u}}_t^D)
\end{equation}
where the fusion kernel $\rm Fuse(\cdot, \cdot)$ is actually a gating layer to combine two representations, we do not give the details here due to space limitation.

To model the long distance dependency issue of document context, we also introduce the self-attention layer to further align the document representation $\textbf{v}_t^D$ against itself, as:
\begin{equation} \label{equ:5}
\begin{gathered}
\beta_{ij}={\rm softmax}(\textbf{v}_i^D\cdot W_s^\top\cdot \textbf{v}_j^D) \\
\tilde{\textbf{v}}_t^D=\sum_{j}\beta_{tj}\cdot \textbf{v}_j^D \\
\textbf{d}_t^D={\rm Fuse}(\textbf{v}_t^D,\tilde{\textbf{v}}_t^D)
\end{gathered}
\end{equation}
where $W_s$ is a trainable bilinear projection matrix. Another fusion kernel is again used to combine the original and self-attentive representations. For all the previous encoding and attention steps,  we process each document independently given the question. Finally, we obtain a question-aware representation $\rm{D}^{D_i}=\{\textbf{d}_t^{D_i}\}$ for each word in each document. 

For the question side, since it is generally short, we directly self-align the question to a vector ${\textbf{r}^Q}$, which is independent from the document, as
\begin{equation} \label{equ:6}
\begin{gathered}
{\gamma_t}=\rm{softmax}( \textbf{w}^\top_q\cdot \textbf{u}_t^Q)\\
{\textbf{r}^Q}=\sum_{t}\gamma_t \cdot \textbf{u}_t^Q
\end{gathered}
\end{equation}
where $\textbf{w}_q$ is a trainable linear weight vector. 

The shared question and document modeling lay the foundation for the subsequent three extraction tasks. Based on the document and question representations $\rm{D}^{D_i}=\{\textbf{d}_t^{D_i}\}$ and ${\textbf{r}^Q}$, we introduce the three joint extraction tasks.

\subsubsection{Document Extraction}
In multi-document MRC, in addition to annotating the answer span, the benchmark datasets generally also annotate which documents are correct for extracting the answer, or it can also be easily obtained given the labeled answer. Therefore, we also introduce an auxiliary document extraction task, to help improve the answer prediction. Compared to the answer span extraction, the document extraction is relatively easier. The aim is to better lay the foundation for the answer prediction and help learn the shared bottom layers.

Firstly, we also self-align the document representation $\rm{D}^{D_i}=\{\textbf{d}_t^{D_i}\}$ for each selected document $\textbf{D}_i$, to obtain a weighted document vector $\textbf{r}^{D_i} $ as:
\begin{equation} \label{equ:7}
\begin{gathered}
{\mu_t}=\rm{softmax}( \textbf{w}^\top_d\cdot \textbf{d}_t^{D_i})\\
{\textbf{r}^{D_i}}=\sum_{t}\mu_t \cdot \textbf{d}_t^{D_i}
\end{gathered}
\end{equation}

Next, the question vector ${\textbf{r}^Q}$ and document vector $\textbf{r}^{D_i}$ are matched in a bilinear function for a relevance score as,
\begin{equation} \label{equ:8}
s^{D_i}={\textbf{r}^Q}\cdot W_{qd}\cdot \textbf{r}^{D_i}
\end{equation}
where $W_{qd}$ is a trainable bilinear projection matrix, which helps to match the two vectors in the same space.

For one question, each selected document $\textbf{D}_i$ has a matching score $s^{D_i}$. We normalize their scores and optimize the following objective function:
\begin{equation} \label{equ:9}
\tilde{s}^{D_i}=1/(1+{\rm exp}^{-s^{D_i}})
\end{equation}\vspace{-2mm}
\begin{equation} \label{equ:9_2}
\mathcal{L}_{DE}=-\frac{1}{K}\sum_{i=1}^{K}[y^{D_i}log\tilde{s}^{D_i}+(1-y^{D_i})log(1-\tilde{s}^{D_i})]
\end{equation}
where $K$ is the number of selected documents. $y^{D_i} \in \{0,1\}$ denotes the label, $y^{D_i}=1$ means document $\textbf{D}_i$ contains one golden answer, otherwise $y^{D_i}=0$.

\subsubsection{Paragraph Extraction}
In general, the golden answer usually comes from one or two paragraphs in each document. We can also annotate the correct paragraphs where the answer is extracted from, by some distant supervision method~\cite{chen2017reading}. Therefore, we introduce a mid-level paragraph extraction task, so that our model can not only distinguish among different documents, but it can also select the relevant paragraphs within each document.  

We first organize each selected document with paragraphs, and follow the same way as in document extraction to calculate a question-paragraph matching score for each paragraph. 
Specifically, for each paragraph in document $\textbf{D}_i$ with $\rm{D}^{D_i}=\{V^{P_{i1}}, \cdots ,V^{P_{iN}}\}$, we first self-align the word-level paragraph representation $\rm V^{P_{ij}}$ to a weighted vector representation $\textbf{r}^{P_{ij}}$ as in Equ.~\ref{equ:7}. Then a bilinear matching function is used between $\textbf{r}^Q$ and $\textbf{r}^{P_{ij}}$ to obtain the corresponding relevance score as:
\begin{equation} \label{equ:10}
s^{P_{ij}}={\textbf{r}^Q}\cdot W_{qp}\cdot \textbf{r}^{P_{ij}}
\end{equation}
where $W_{qp}$ is the trainable bilinear projection matrix between question and paragraph.

For one document, each paragraph $\textbf{P}_{ij}$ in the document has a matching score $s^{P_{ij}}$. We normalize the scores among each document and obtain $\tilde{s}^{P_{ij}}$ as in Equ.~\ref{equ:9}. In this sub-task, we optimize the average cross-entropy loss among all the selected documents and paragraphs as:
\begin{equation} \label{equ:11}
\mathcal{L}_{PE}=-\frac{1}{K}\frac{1}{N}\sum_{i=1}^{K}\sum_{j=1}^{N}[y^{P_{ij}}log\tilde{s}^{P_{ij}}+(1-y^{P_{ij}})log(1-\tilde{s}^{P_{ij}})]
\end{equation}
where $N$ is the number of remaining paragraphs for each document. $y^{P_{ij}} \in \{0,1\}$ denotes the paragraph-level label for the $j^{th}$ paragraph in $i^{th}$ document. 

\subsubsection{Answer Span Extraction}
The ultimate goal is to predict a correct answer, where the afore-mentioned document extraction and paragraph extraction actually act as two auxiliary tasks, so that the shallow semantic representations can be better learnt. In this stage, we aim to combine all the available information to accurately extract the answer from all the selected documents at a span level. To make the document representation aware of information in different documents and enable a direct comparison across different documents, we concatenate all the selected documents together and introduce a muilti-document shared LSTM layer for contextual modeling as:
\begin{equation} \label{equ:12}
\textbf{g}_t^D=\rm{BiLSTM}(\textbf{g}_{t-1}^D, [{\textbf{d}_t^D};{\textbf{r}^Q};\textbf{f}])
\end{equation}
where $\textbf{f}$ is a manual feature vector including the popular features such as whether each document word occurs in the question words and whether the word is a sentence ending separator. Here we also concatenate the question vector  $\textbf{r}^Q$ to each word representation $\textbf{d}_t^D$ of the document for better modeling the interaction. 

Since all the words from different documents will be passed to the shared LSTM layer, the sequence order is thus very important. We follow the document ranking order obtained via the document ranking function in document retrieval module, as is shown in top of Figure~\ref{fig:cascade}. In this way, we expect that the answer prediction model can also bear the ranking relevance in document retrieval module in mind and it shows good performance in our experiment. 

Finally, the pointer network~\cite{wang2016machine} is used to predict the start and end position of the answer with the probabilities $\alpha_t^1$ and $\alpha_t^2$,
and the answer extraction model can be trained by minimizing the negative log probabilities of the true start and end indices:
\begin{equation} \label{equ:12_1}
\alpha_t={\rm exp}(\textbf{w}_a^\top \textbf{g}_t^D)/\sum_{j=1}^{|D_w|}{\rm exp}(\textbf{w}_a^\top \textbf{g}_j^D)
\end{equation}
\begin{equation} \label{equ:13}
\mathcal{L}_{AE}=-\frac{1}{M}\sum_{i=1}^{M}(log\alpha^1_{y^1_i}+log\alpha^2_{y^2_i})
\end{equation}
where $\textbf{w}_a$ is a trainable vector, $|D_w|$ is the total number of words. $M$ is the number of question samples, $y^1_i$, $y^2_i$ are the golden start and end positions across the entire documents. 

\subsubsection{Joint Training and Prediction}
According to the design, the three extraction tasks share the same embedding, encoding and matching layers. Therefore, we propose to train them together as multi-task learning. The joint objective function is formulated as follows:
\begin{equation} \label{equ:14}
\mathcal{L}=\mathcal{L}_{AE}+\lambda_1\mathcal{L}_{DE}+\lambda_2\mathcal{L}_{PE}
\end{equation}
where $\lambda_1$ and $\lambda_2$ are two hyper-parameters that control the weights of those tasks.

To keep the training process stable, we adopt a coarse-to-fine joint training strategy and progressively finetune one upper task with the joint loss. Specifically, we first train the downside document extraction and paragraph extraction tasks to obtain an initial shallow representation, and then jointly train the three tasks with Equ.\ref{equ:14} based on it. Besides, when training a new upper task, we follow the method in~\cite{hashimoto2016joint} and introduce a successive regularization term on the shared parameters, as:
\begin{equation} \label{equ:15}
\mathcal{L}=\mathcal{L}+\delta ||\theta_s-{\theta}'_s||^{2}
\end{equation}
where $\theta_s$, ${\theta}'_s$ are the shared parameters at successive training stages. In this way, we can restrain the joint training process so that the shared parameters will not change so much. 

When predicting the final answer, we take the document matching score, paragraph matching score and answer span score into consideration and choose the answer with the highest prediction score, given as:
\begin{equation} \label{equ:16}
s=(\alpha_k^1\cdot \alpha_k^2)\cdot \tilde{s}^{D_i} \cdot \tilde{s}^{P_{ij}}
\end{equation}

\section{Experiments}

This section presents the experimental methodology. We first verify the effectiveness of our model on two benchmark datasets: TriviaQA~\cite{joshi2017triviaqa} and DuReader~\cite{he2017dureader}. Then we test our model in  operational online environment, which can stably and effectively serve different scenarios promptly. 


\subsection{Datasets}

\subsubsection{Off-line Benchmark Dataset}
We choose the TriviaQA Web and DuReader benchmark datasets to test our method, since both of them are multi-document MRC datasets which is more realistic and challenging. 

TriviaQA is a recently released large-scale multi-document MRC datasets, which consists of 650K context-query-answer triples. There are 95K distinct question-answer pairs, which are 
authored by Trivia enthusiasts, with 6 evidence documents (context) per question on average, which are generated from either Wikipedia or Web search. In this paper, we focus on the TriviaQA Web dataset, which contains more context data for each question. 

DuReader is so far the largest Chinese MRC dataset, which contains 200K questions, 1M documents and more than  420K human-summarized answers. All the questions and documents are extracted from real data, by the largest Chinese search engine Baidu. The average length of the documents is 396.0 words, and on average each question has 5 evidence documents, each document has about 7 paragraphs.

\subsubsection{On-line Environment}
We also apply our model to the AliMe Chatbot system, which is an intelligent online assistant designed for creating an innovative online shopping experience in e-commerce. Currently, it serves millions of customer questions per day. We test our model in two practical scenarios, i.e., e-commerce promotion and tax policy reading. E-commerce promotion scenario is about consulting instructions on shopping games and sales promotion, which mostly involves with a short document with no more than 500 words. Tax policy scenario is about reading tax policy articles, which can be viewed as a multi-document MRC task. The length of the article is much longer, which consist of many sections and paragraphs.  

\subsection{Implementation Details}
For the cascade ranking functions, the number of selected documents $K$ and paragraphs $N$ are the key factors to balance the effectiveness and efficiency trade-off. We choose $K=4$ and $N=2$ for the good performance when evaluating on the dev set. Since the TriviaQA documents often contain many small paragraphs, we also restructure the documents by merging consecutive paragraphs to a maximum size of 600 words for each paragraph as in~\cite{clark2017simple}. The detailed analysis will be given and discussed in the next section.

For the multi-task deep attention framework, we adopt the Adam optimizer for training, with a mini-batch size of 32 and initial learning rate of 0.0005. We use the GloVe 300 dimensional word embeddings in TriviaQA and train a word2vec word embeddings with the whole DuReader corpus for DuReader. The word embeddings are fixed during training. The hidden size of LSTM is set as 150 for TriviaQA and 128 for DuReader. The task-specific hyper-parameters $\lambda_1$ and $\lambda_2$ in Equ.~\ref{equ:14} are set as $\lambda_1=\lambda_2=0.5$. Regularization parameter $\delta$ in Equ.~\ref{equ:15} is set as a small value of 0.01. All models are trained on Nvidia Tesla M40 GPU with Cudnn LSTM cell in Tensorflow 1.3. 


\subsection{Off-line Evaluation}

\subsubsection{Main Results}
The results of our single deep cascade model~\footnote{\small{We only submit the single model without any model ensemble.}} on TriviaQA Web and DuReader 1.0 are summarized in Table~\ref{tab:1} and Table~\ref{tab:2}, respectively. We can see that by adopting the deep cascade learning framework, the proposed model outperforms the previous state-of-the-art methods by an evident margin on both datasets, which validates the effectiveness of the proposed method in addressing the challenging multi-document MRC task.


\begin{table}[t]
\scriptsize
\centering
\caption{\label{tab:1} Performance of our method and competing models on the TriviaQA Web leaderboard.}
\begin{tabular}{lcc}
\hline
     & \textbf{Full} & \textbf{Verified} \\
\hline
   \textbf{Model} & \textbf{EM} / \textbf{F1} & \textbf{EM} / \textbf{F1} \\
\hline
    BiDAF Baseline \cite{joshi2017triviaqa} & 40.74 / 47.05 &  49.54 / 55.80 \\
    Smarnet \cite{chen2017smarnet} & 40.87 / 47.09 & 51.11 / 55.98 \\
    M-Reader \cite{hu2017reinforced} & 46.65 / 52.89 & 56.96 / 61.48 \\
    Re-Ranker \cite{wang2017evidence} &  63.04 / 68.53 &  69.70 / 74.57 \\
    S-Norm \cite{clark2017simple} &  66.37 / 71.32 &  79.97 / 83.70 \\
    Weissenborn \cite{weissenborn2017dynamic}  &  67.46 / 72.80 &  77.63 / 82.01 \\
\hline
    \textbf{Our-Single} & \textbf{68.65} / \textbf{73.07}  & \textbf{82.44} / \textbf{85.35} \\
\hline
\end{tabular} \vspace{-3mm}
\end{table} 

\begin{table}[t]
\scriptsize
\centering
\caption{\label{tab:2} Performance on the DuReader 1.0 test set.}
\begin{tabular}{lcc}
\hline
   \textbf{Model} & \textbf{BLEU-4}  & \textbf{ROUGE-L} \\
\hline
    Match-LSTM \cite{wang2016machine} & 31.8 &  39.0 \\
    BiDAF \cite{seo2016bidirectional} & 31.9 & 39.2 \\
    PR + BiDAF \cite{wang2018multi} &  37.55 &  41.81 \\
    Cross-Passage Verify \cite{wang2018multi} &  40.97 &  44.18 \\
    R-net \cite{wang2017gated} & 44.88 & 47.71 \\
    \textbf{Our-Single} &  \textbf{49.39} &  \textbf{50.71} \\
\hline
    Human Performance & 56.1  & 57.4 \\
\hline
\end{tabular} \vspace{-3mm}
\end{table}

\subsubsection{Ablation Study}
To get better insight into our model architecture, we conduct an in-depth ablation study on the development set of DuReader and TriviaQA, which is shown in Table~\ref{tab:3}. The main goal is to validate the effectiveness of the critical components in our architecture including the manual features and multi-document shared LSTM in the pure answer span extraction task, the cascade document and paragraph ranking functions for pruning irrelevant document content and the adoption of multi-task learning strategy.

From the results, we can see that: 1) the shared LSTM plays an important role in answer extraction among multiple documents, the benefit lies in two parts: a) it helps to normalize the content probability score from multiple documents so that the answers extracted from different documents can be directly compared; b) it can keep the ranking order from document ranking component in mind, which may serve as an additional signal when predicting the best answer. By incorporating the manual features, the performance can be further improved slightly. 2) Both the preliminary cascade ranking and multi-task answer extraction strategy are vital for the final performance, which serve as a good trade-off between the pure pipeline method and fully joint learning method. By removing the rich irrelevant noisy data in the cascade document and paragraph ranking stage, the downside MRC model can better extract the       answer from the more relevant content data. Jointly training the three extraction tasks can provide great benefits, which shows that the three tasks are actually closely related and can boost each other with shared representations at bottom layers.

\begin{table}[t]
    \scriptsize
    \centering
    \caption{\label{tab:3} Ablation study on model components. }
    \begin{tabular}{l|cc|cc}
    \hline
       \multirow{2}*{\textbf{\quad \quad \quad Model}} & \multicolumn{2}{c|}{\textbf{DuReader}}  & \multicolumn{2}{c}{\textbf{TriviaQA}} \\ 
       \cline{2-5}
       & \textbf{Bleu-4 score}  & \textbf{$\Delta$} & \textbf{F1} & \textbf{$\Delta$} \\
    \hline
        \bf{Complete Model} & \textbf{50.8} & -- & \textbf{73.8} & --\\
        w/o Manual Features & 49.8 & -1.0 & 73.0 & -0.8\\
        w/o Shared LSTM & 48.7 & -2.1 & 70.5 & -3.3 \\
        w/o Cascade Ranking & 47.0 & -3.8 & 71.1 & -2.7\\
        w/o Multi-task Learning & 48.5 & -2.3 & 70.9  & -2.9\\
        \bf{Boundary Baseline} & 41.0 & -9.8 & 61.5 & -12.3\\
    \hline
    \end{tabular} \vspace{-3mm}
    \end{table} 
    
\begin{table}[t]
    \scriptsize
    \centering
    \caption{\label{tab:4} Effectiveness and efficiency w.r.t document and paragraph selection number on DuReader development set (Efficiency is indicated by time cost at prediction stage).}
    \begin{tabular}{cccc}
    \hline
       \textbf{Document No.} & \textbf{Paragraph No.}  & \textbf{Time cost(s) / batch}  & \textbf{Bleu-4 score}\\
    \hline
        \multirow{3}*{1} & 1 &  0.42 & 32.1 \\
        ~ & 2 &  0.53 & 35.4 \\
        ~ & 3 &  0.69 & 36.0 \\
    \hline
        \multirow{3}*{2} & 1 &  0.56 & 40.5 \\
        ~ & 2 &  0.89 & 44.8 \\
        ~ & 3 &  1.14 & 44.2 \\
    \hline
        \multirow{3}*{3} & 1 &  0.71 & 48.1 \\
        ~ & 2 &  1.09 & 49.5 \\
        ~ & 3 &  1.36 & 49.0 \\
    \hline
        \multirow{3}*{4} & 1 &  0.88 & 49.6 \\
        ~ & 2 &  1.39 & \textbf{50.8} \\
        ~ & 3 &  1.75 & 50.0 \\
    \hline
        \multirow{3}*{5} & 1 &  0.98 & 49.6 \\
        ~ & 2 &  1.70 & 50.0 \\
        ~ & 3 &  2.03 & 48.8 \\
    \hline
    \end{tabular} \vspace{-3mm}
    \end{table} 

\subsubsection{Effectiveness v.s. Efficiency Trade-off}
Now we further examine how the performance of our model changes with respect to the number of selected documents and paragraphs in cascade ranking stage, which is the key factor to control the effectiveness and efficiency trade-off. The result on DuReader development set is presented in Table~\ref{tab:4}. We can see that: 1) By properly taking more documents or paragraphs into consideration, the performance of the model gradually increases when it reaches 4 documents and 2 paragraphs, and then the performance decreases slightly which may be due to that much noisy data is introduced.  2) The time cost can be largely reduced by removing more irrelevant documents and paragraphs in the cascade ranking stage, while keeping the performance not change that much. For example, for the best setting at 4 documents and 2 paragraphs, if we instead only keep the top-1 paragraph for each document, the time cost will be reduced by 36.7\%, while the performance only decreases about 2.4\%. As a result, we can adaptively change our model to meet the practical situation and we choose 4 documents and 2 paragraphs in our off-line experiment where effectiveness is most emphasized.


\subsubsection{Advantage of Multi-task Learning}
Next, we also analyze the benefits brought in via the adoption of the multi-task learning strategy in detail. The performance of jointly training the answer extraction module with different auxiliary tasks on DuReader development set is shown in Table~\ref{tab:5}. We can see that by incorporating the auxiliary document extraction or paragraph extraction task in the joint learning framework, the performance can always improve which again shows the advantage of introducing auxiliary tasks for helping to learn shared bottom representations. Besides, the performance gain by adding document extraction task is larger, which may be due to that it can better lay the foundation of the model with that information from different documents can be distinguished. 


\begin{table}[t]
    \scriptsize
    \centering
    \caption{\label{tab:5} Performance with different extraction tasks.}
    \begin{tabular}{lcc}
    \hline
       \textbf{Task} & \textbf{Bleu-4 score}  & \textbf{$\Delta$} \\
    \hline
    	Pure Answer Span Extraction & 48.5 & -- \\
	+ Document Extraction & 49.7 & +1.2 \\
	+ Paragraph Extraction & 49.2 & +0.7 \\
	+ Document \& Paragraph Extraction & 50.8 & +2.3 \\
    \hline
    \end{tabular} \vspace{-3mm}
    \end{table}


\subsection{On-line Evaluation}


\subsubsection{Results on E-commerce and Tax data}

\begin{table}[t]
\scriptsize
\centering
\caption{\label{tab:6} Performance and response time (RT) in two real-world online scenarios.}
\begin{tabular}{lcc}
\hline
        & \textbf{Tax} & \textbf{E-Commerce} \\
\hline
    \textbf{Model} & \textbf{F1} / \textbf{RT} & \textbf{F1} / \textbf{RT} \\
\hline
    BiDAF \cite{joshi2017triviaqa} & 40 / 130ms &  63 / 65ms \\
    DrQA \cite{chen2017reading} & 46 / 122ms & 67 / 61ms \\
    Our-Single (w/o Cascade Ranking) & 55 / 138ms &  71 / 70ms \\
    \textbf{Our-Single (K=3, N=1)} & \textbf{76.5} / \textbf{45ms} &  \textbf{73} / \textbf{38ms} \\
\hline
\end{tabular} \vspace{-3mm}
\end{table} 

We also test the effectiveness and efficiency of our model in two practical scenarios, E-commerce and tax policy reading, where real-time responses are expected and a large number of customers are being served simultaneously. The comparative result is shown in Table~\ref{tab:6}. We can see that by introducing the cascade ranking stage and keeping the selected number properly, our method can serve the requests with a much higher speed of less than 50ms, especially for tax scenario where the improvement is about 3 times. Besides, the performance with respect to F1 score is also largely improved with the proposed multi-document MRC model, which demonstrates the effectiveness of our method for removing the rich irrelevant noisy content in our online scenario.


\subsubsection{Results on Different Document Lengths}
\begin{figure}
\centering
\includegraphics[width=0.499\textwidth]{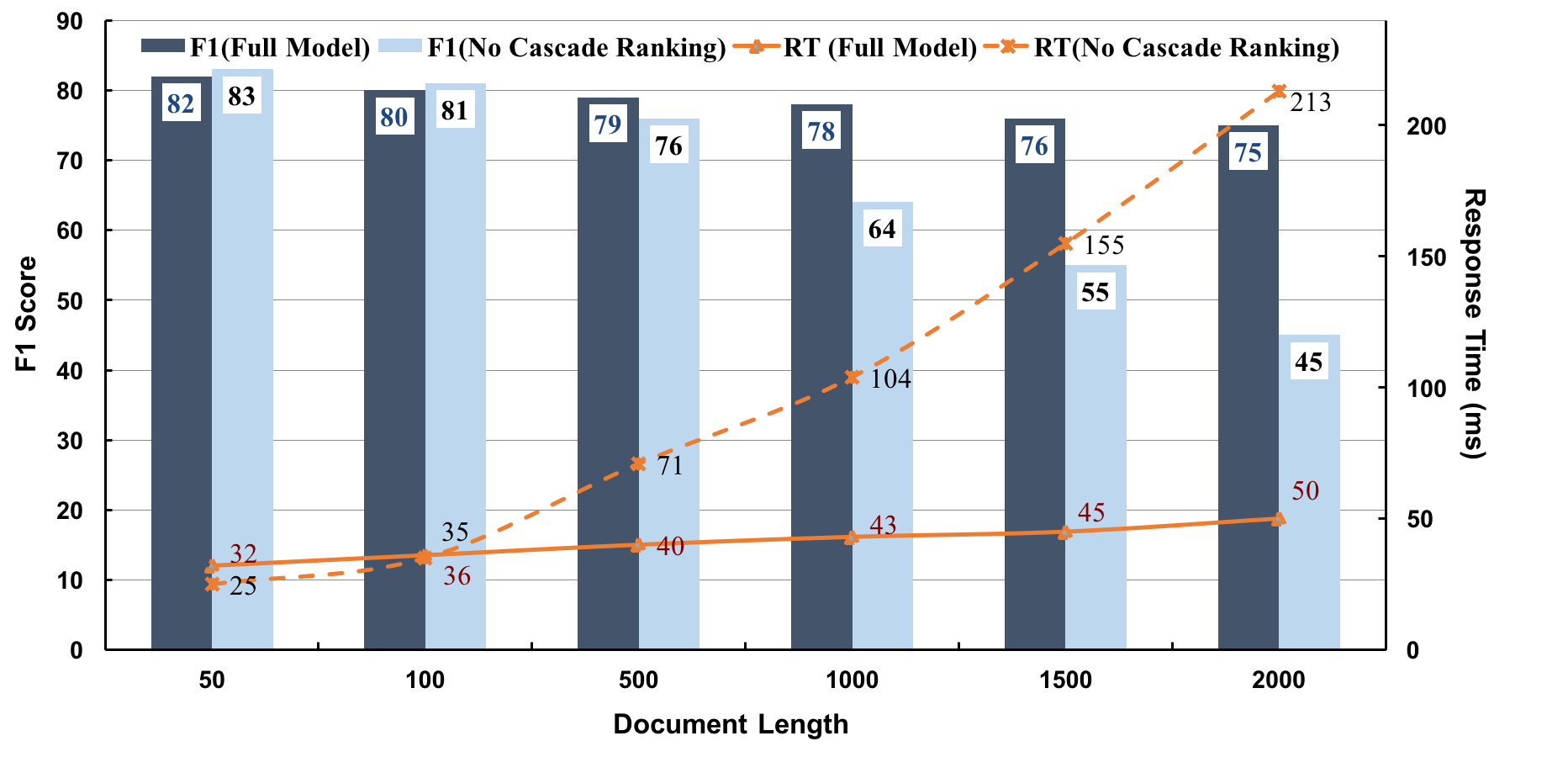} \vspace{-4mm}
\caption{F1 score and average response time w.r.t different document lengths.} \vspace{-2mm}
\label{fig:retrieval_efficiency} \vspace{-3mm}
\end{figure}

We further examine how the F1 score and response time change on tax scenario when processing documents with different lengths, ranging from 50 to 2000 words. The result is shown in Figure~\ref{fig:retrieval_efficiency}. We can see that without incorporating with the cascade ranking module, the answer extraction module performs rather poorly both in effectiveness and efficiency as the document length increases. In particular, when the document length exceeds 1,000 the total response time increases 3 to 6 times, while for our full cascade model only 15ms more are needed. 

\section{Conclusion}

In this paper, we propose a novel deep cascade learning framework to balance the effectiveness and efficiency in the more realistic multi-document MRC. We design three cascade modules, which can eliminate irrelevant document content in the earlier stages with simple features and models, and discern more relevant answers at later stages. The experiment results show that our method can achieve state-of-the-art performance on two large-scale benchmark datasets. Besides, the proposed method has also been effectively and efficiently applied in our online system.


\bibliography{aaai2019}
\bibliographystyle{aaai}

\end{document}